\documentclass[runningheads,a4paper]{llncs}

\usepackage{amssymb}
\usepackage{amsmath}
\usepackage{graphicx}
\usepackage{diagbox}
\usepackage{subcaption}
\usepackage{url}
\usepackage[misc]{ifsym}
\newcommand{\keywords}[1]{\par\addvspace\baselineskip
\noindent\keywordname\enspace\ignorespaces#1}
\newcommand{\al}{et al. }
\spnewtheorem{hypo}[theorem]{Hypothesis}{\bfseries}{\itshape}
\spnewtheorem{principle}[theorem]{Principle}{\bfseries}{\itshape}
\begin{document}

\mainmatter 

\title{Routine Modeling\\with Time Series Metric Learning}
\titlerunning{}

\author{Paul Compagnon\inst{1, 2}, Gr\'{e}goire Lefebvre\inst{1}, \\ Stefan Duffner\inst{2} \and Christophe Garcia\inst{2}}

\institute{Orange Labs,
Grenoble, France\\ \{paul.compagnon, gregoire.lefebvre\}@orange.com 
\and LIRIS, UMR 5205 CNRS INSA-Lyon, France\\
\{stefan.duffner, christophe.garcia\}@liris.cnrs.fr
}

\toctitle{Lecture Notes in Computer Science}
\tocauthor{Authors' Instructions}
\maketitle

\setcounter{footnote}{0}
\begin{abstract}
Traditionally, the automatic recognition of human activities is performed with supervised learning algorithms on limited sets of specific activities. This work proposes to recognize recurrent activity patterns, called routines, instead of precisely defined activities. The modeling of routines is defined as a metric learning problem, and an architecture, called SS2S, based on sequence-to-sequence models is proposed to learn a distance between time series. This approach only relies on inertial data and is thus non intrusive and preserves privacy. Experimental results show that a clustering algorithm provided with the learned distance is able to recover daily routines.
\end{abstract}

\section{Introduction}
Human Activity Recognition (HAR) is a key part of several intelligent systems interacting with humans: smart home services \cite{cumin2017human}, actigraphy and telemedecine, sport applications \cite{avci2010activity}, etc. It is particularly useful for developing eHealth services and monitoring a person in its everyday life. It has been so far mainly performed in supervised contexts with data annotated by experts or with the help of video recordings \cite{chatzaki2016human}.
Not only is this approach time consuming, but it also restricts the number of activities that can be recognized. It is associated with scripted datasets where subjects are asked to perform sequences of predefined tasks. This approach is thus unrealistic and difficult to set up for real environments where people do a vast variety of specific activities everyday and can diverge from a pre-established behavior in many different ways (e.g., falls, accidents, contingencies of life, etc.). 
Besides, most people present some kind of habitual behavior, called \textit{routines} in this paper: the time they go to sleep, morning ritual before going to work, meal times, etc. Results from behavioral psychology show that habits are hard and long to form but also hard to break when well installed \cite{lally2010habits}. 
From a data-driven perspective, Gonzalez \al \cite{gonzalez2008understanding} observed the high regularity of human trajectories thanks to localization data and show that ``humans follow simple reproducible patterns''. Routines produce distinguishable patterns in the data which, if not identifiable semantically, could be retrieved over time and so produce a relevant signature of the daily life of a person. 
In this paper, we advocate for the modeling of such routines instead of activity recognition, and we propose a machine learning model able to identify routines in the daily life of a person. We want this system to be unintrusive and to respect people's privacy and therefore to rely only on inertial data that can be gathered by a mobile phone or a smart watch. Moreover, routines do not need to be semantically characterized, and the model does not have to use any activity labels. The daily routines of a person may present characteristics of almost-periodic functions, periodic similarity, regarding a certain metric which we propose to learn. To do so, we adapted the siamese neural network architecture proposed by Bromley \al\cite{bromley1994signature} to learn a distance from pairs of sequences and propose experiments to evaluate the quality of the learned metric on the problem of routine modeling. The contributions of this paper are threefold: 

\begin{enumerate}
\item a formulation of routine modeling as a metric learning problem by defining routines as almost-periodic functions, 
\item an architecture to jointly learn a representation and a metric for time series using siamese sequence-to-sequence models and an improvement of the loss functions to minimize,
\item results showing that the proposed architecture is effectively able to recover human routines from inertial data without using any activity labels.
\end{enumerate}
The remainder of the paper is organized as follows. Section \ref{contrib} is dedicated to routine modeling definition. Section \ref{rw} gives an overview of time series metrics. The proposed approach to recognize routines is presented in Section \ref{srnn} and Section \ref{exp} presents experimental protocols and results. Finally, conclusions and perspectives are drawn in the last section. 

\section{Routine Modeling}
\label{contrib}

A routine can be seen as a recurrent behavior of an individual's daily life. For example, a person roughly does the same thing in the same order when waking up or going to work. These sequences of activities should produce distinguishable patterns in the data and can thus be used to monitor the life of an individual without knowing what he or she is doing exactly. The purpose of this work is to design an intelligent system which is able to recognize routines. To tackle routines with machine learning, we propose a starting principle similar to the one used in natural language processing: \textit{similar words appear in similar contexts}. The context surrounding a word designates the previous and following words of the sentence, for example. The context of a routine corresponds here to the moment of the day or the week, etc. it generally happens.
\begin{principle}
Similar routines occur at similar moments, almost periodically.
\end{principle} 
\setcounter{theorem}{0}
From this principle, we seek now to propose a mathematical formulation of routines which would include the notions of periodicity and similarity. The almost periodic functions defined by Bohr \cite{bohr1925theorie} show similar properties: 
\begin{definition}
Let $f: \mathbb{R} \rightarrow \mathbb{C}$ be a continuous function. $f$ is an almost-periodic function with respect to the uniform norm if $\forall \epsilon > 0$, $\exists T > 0$ called an $\epsilon$-almost period of $f$ such as: 
\begin{equation}
\text{sup}\ |f(t+T) - f(t)| \leqslant \epsilon.
\end{equation}

\end{definition}
 
Obviously, the practical issue of routine modeling presents several divergences from this canonical definition: data are discrete time series and the periodicity of activities cannot be evaluated point-wise. Nevertheless, it is possible to adapt it to our problem. Let $S: \mathbb{N} \rightarrow \mathbb{R}^n$ be an ordered discrete sequence of vectors of dimension $n$. If the frequency of $S$ is sufficiently high, it is possible to get a continuous approximation of it, by interpolation for example. We now consider a function $f_S$ of the following form with a fixed interval length $l$: 
\begin{equation}
\begin{split}
f_S: \ \mathbb{R}_{+} & \rightarrow \mathbb{R}^{n \times l} \\
  t & \mapsto [S(t):S(t+l)[, 
\end{split}
\end{equation}
where $[S(t):S(t+l)[$ is the set of vectors between $S(t)$ and $S(t+l)$ sampled at a certain frequency from the continuous approximation. $l$ is typically one or several hours: a sufficiently long period of time to absorb the little changes from one day to another (e.g., waking up a little earlier or later, etc.). The objective is to define almost-periodicity with respect to a distance $d$ between sequences, such that $\forall \epsilon > 0$, $\exists T > 0$: 
\begin{equation}
\label{equ3}
d(f_{S}(t), f_{S}(t+T)) \leqslant \epsilon .
\end{equation}
The parameter $T$ can be a day, a week or a sufficiently long period of time to observe repetitions of behavior. The metric $d$ must be sufficiently flexible to handle the high variability of activities which can be similar but somewhat different in their execution while exhibiting a similar pattern. We therefore postulate that $d$ may be learned for a specific user from its data and we will now show that $f_S$ respects the condition established in Eq. \eqref{equ3} with respect to $d$.
To learn $d$ if pairs of similar and dissimilar sequences are known, a Recurrent Neural Network (RNN) encoder parametrized by $W$, called $G_W$, can encode the sequences into vector representations and the contrastive loss \cite{hadsell2006dimensionality} can be used to learn the metric from pairs of sequence encodings:
\begin{equation}
\label{hadsell}
L(W, Y_1, Y_2, y) = (1-y)\frac{1}{2}d(Y_1, Y_2)^2 + y \frac{1}{2} \text{max}(0, m- d(Y_1, Y_2))^2,
\end{equation} 
where $y$ is equal to zero or one depending if the sequences are respectively similar or not, $Y_1$ and $Y_2$ are the last output of the RNN for both sequences and $m >0$ a margin that defines the minimal distance between dissimilar samples. Several justifications arise for the use of a margin in metric learning. It is necessary to prevent flat energy surface, according to energy-based learning theory \cite{lecun2006tutorial}, a situation where the energy is low for every input/output associations, not only those in the training set. It also insures that metric learning models are robust to noise \cite{weinberger2009distance}. As the learning process aims to minimize the distances between similar sequences which are, by definition, shifted by a period $T$, we get, for a fixed $T > 0$ and $\forall t \in \mathbb{R}_+$: 
\begin{equation}
\label{equ5}
d(G_W(f_{S}(t)), G_W(f_{S}(t+T))) \leqslant m.
\end{equation}
The margin $m$ can be chosen as close to zero as possible and thus Eq. \eqref{equ5} identifies itself with Eq. \eqref{equ3}. In practice, this optimization is only possible up to some point, depending on the model and the data. This argumentation suggests the interest of modeling routines with metric learning as, in this case, the main property of almost-periodic functions is fulfilled. 

\section{Related Work}
\label{rw}

The traditional approach to compute distances between sequences (or time series, or trajectories) is to perform Dynamic Time Warping (DTW) \cite{sakoe1978dynamic} which was introduced in 1978. Since then, several improvements of the algorithm have been published, notably a fast version by Salvador \al \cite{salvador2007toward}. DTW is  considered one of the best metric to use for sequence classification \cite{xi2006fast} combined with $k$-nearest neighbors. Recently, Abid \al \cite{abid2018autowarp} proposed a neural network architecture to learn the parameters of a warping distance accordingly to the euclidean distances in a projection space. However, DTW, as other shaped-based distances \cite{esling2012time}, is only able to retrieve local similarities when time series have a relatively small length and are just shifted or not well aligned.

Similar routines could present different data profiles which would necessitate a more complex and global notion of similarity. This justifies the extraction of high-level features to produce a vector representation of the structure and the semantics of the data \cite{lin2009finding}. Traditional metrics can be used to compare vector representations: Euclidean, cosine or Mahalanobis. These vectors can be build with features extracted by various methods such as discrete Fourier and Wavelet transforms, signal processing, singular value decomposition or Hidden Markov Models (HMM) \cite{aghabozorgi2015time}. HMM belong to a category of approaches which suppose the existence of an underlying model which has produced the data; other examples include AutoRegressive-Moving-Average (ARMA) or multivariate extensions (VARIMA), Markov chains, etc. In this case, similarity can be assess by comparing model parameters. More theoretical approaches based on the study of the spectral properties of these models have also been proposed in \cite{kalpakis2001distance,martin2000metric}. The problem with these approaches is that it is difficult to select relevant features and/or to chose an accurate model and parameters for a given task. It would be better if an appropriate representation of the data could automatically be extracted accordingly to the problem, by a Neural Network (NN) for example. 

Besides, Bromley \al \cite{bromley1994signature} proposed a Siamese Neural Network (SNN) architecture to learn a metric. They have since then been used for many applications with feedforward or convolutionnal NN such as person reidentification \cite{yi2014deep}, gesture recognition \cite{berlemont2018class}, object tracking \cite{bertinetto2016fully}, etc. RNN and particularly Long-Short Term Memory (LSTM) NN \cite{hochreiter1997long} are well-adapted to work with long sequential data as they are able to deal with long-term dependencies. M\"{u}ller \al \cite{mueller2016siamese} used a siamese recurrent architecture to learn sentence similarity by encoding sequence of word vectors previously extracted belonging to the same sentence.

In the following section, we propose a novel Siamese Sequence to Sequence (SS2S) neural network architecture to learn to model routines without label supervision. The model effectively combines automatic feature extraction and a similarity metric by jointly learning a robust projection of time series in a metric space. This approach is able to deal with long sequences by using LSTM networks and do not necessitate to choose a model to fit or features to extract. 

\section{Siamese Sequence to Sequence Model}
\label{srnn}

\begin{figure}[t]
\centering
\includegraphics[width=\textwidth]{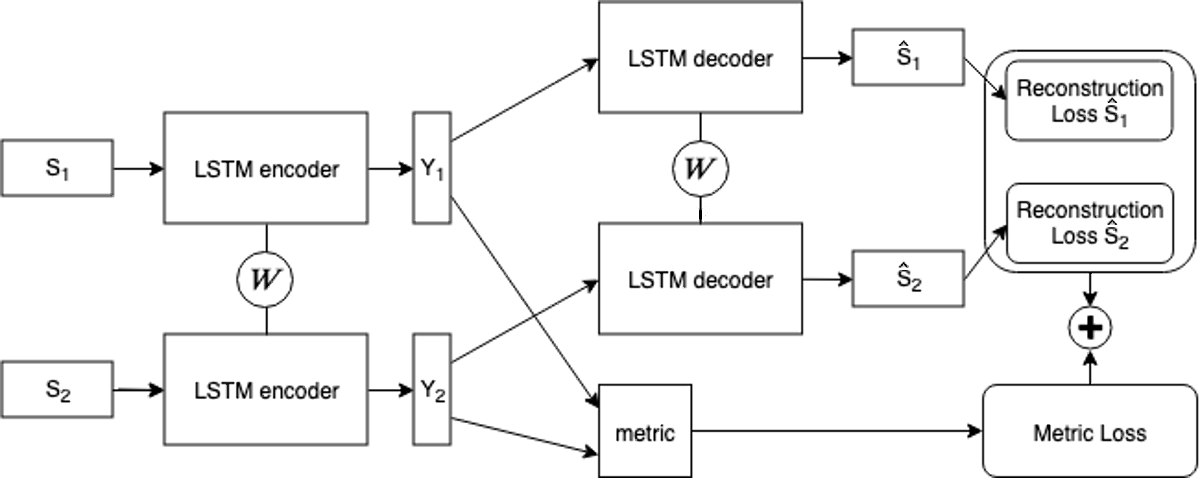}
\caption{Proposed SS2S architecture.}
\label{archi}
\end{figure}

\subsection{Feature Extraction Approach}
The time series data obtained from inertial sensors may be very noisy and certainly vary for the same general activity (e.g., cooking). Robust feature representations of time series should therefore be learned before learning a metric. We thus propose (Fig. \ref{archi}) to map each sequence to a vector using a Sequence to Sequence model \cite{abid2018autowarp,cho2014learning,sutskever2014sequence}. The sequence is given as input to the first LSTM network (the encoder) to produce an output sequence, the last output vector is considered as the learned representation. This representation is then given to the second LSTM (the decoder) which tries to reconstruct the input sequence. Typically, an autoencoder is trained to reconstruct the original sequence with the Mean Squared Error (MSE):
\begin{equation}
\text{MSE}(S, \hat{S}) = \frac{1}{l} \sum_{t=0}^{l-1}(S(t) - \hat{S}(t))^2,
\end{equation}
where $S$ is the sequence and $\hat{S}$ the output sequence produced by the autoencoder from the vector. Similarly, we propose a new Reconstruction Loss (RL) based on cosine similarity, the Cosine Reconstruction Loss (CRL):
\begin{equation}
\text{CRL}(S, \hat{S}) = l - \sum_{t= 0}^{l-1} \text{cos}(S(t), \hat{S}(t)).
\end{equation} 
CRL is close to 0 if the cosine similarity between each pair of vectors is close to one when the vectors are collinear.

\subsection{Metric Learning} 
Our architecture is a siamese network \cite{bromley1994signature}, that is to say it is constituted of two subnetworks sharing the same parameters $W$ (see Fig. \ref{archi}). It takes pairs of similar or dissimilar sequences as input constituted with what is called \textit{equivalence constraints}. The objective of our architecture is therefore to learn a metric which makes close similar elements and separates the dissimilar ones in the projection space. Three metric forms can generally be used: Euclidean, cosine or Mahalanobis \cite{hadsell2006dimensionality,yi2014deep,faraki2018large}. The first two are not parametric and only a projection is learned. Learning a Mahalanobis-like metric implies not only learning the projection but also the matrix which will be used to compute the metric. One different Metric Loss (MeL) is proposed to learn each metric form. $Y_1$ and $Y_2$ are the representations learned by the autoencoder from the inputs of the siamese network. The first is the contrastive loss \cite{hadsell2006dimensionality} (see Eq. \eqref{hadsell}) to learn an euclidean distance. The second is a cosine loss to learn a cosine distance:
\begin{equation}
L(W, Y_1, Y_2, y) = \left\{ 
							\begin{array}{c c c}
							1 - \text{cos}(Y_1, Y_2), & \text{if}\; y&= 1 \\
							\text{max}(0, \text{cos}(Y_1, Y_2) - m), & \text{if}\; y&= -1.
							\end{array}
							\right.
\end{equation}
Finally, Mahalanobis metric learning can be performed with the KISSME algorithm \cite{koestinger2012large} which can be integrated into a NN \cite{faraki2018large}. This algorithm aims to maximize the dissimilarity log-likelihood of dissimilar pairs and conversely for similar pairs. The model learns a mapping under the form of a matrix $W$ and an associated metric matrix $M$ of the dimension of the projection space. $W$ is integrated into the network as a linear layer (just after the recurrent encoding layers in SS2S) trained with backpropagation while $M$ is learned in a closed-form manner and updated after a fix number of epochs with the following formula: 
\begin{equation}
M = \text{Proj}((W^{T} \Sigma_S W)^{-1} - (W^{T} \Sigma_D W)^{-1}).
\end{equation}
$\Sigma_S$ and $\Sigma_D$ are the covariance matrices of similar and dissimilar elements in the projection space and $\text{Proj}$ is the projection onto the positive semi definite cone. We propose a modified version of the KISSME loss proposed in \cite{faraki2018large} which we found was easier to train based on the contrastive loss (Eq. \eqref{hadsell}): 
\begin{equation}
\begin{split}
L(W, Y_1, Y_2, y) = &(1-y)\frac{1}{2}(Y_1 - Y_2)M(Y_1 - Y_2)^T \\
&+ y \frac{1}{2} \text{max}(0, m- (Y_1 - Y_2)M(Y_1 - Y_2)^T).
\end{split}
\end{equation}

\subsection{Training Process}
Two training processes can be considered for this architecture. Train the autoencoder and then ``freeze'' the network parameters to learn the metric if it is parametric. Or, add the metric loss to the reconstruction loss and learn jointly both tasks. In this case, several difficulties could appear. Both losses must have similar magnitudes to have similar influences on the training process. The interaction between the two must also be considered. Both tasks could have eventually divergent or not completely compatible objectives. Indeed, we proposed the CRL with the \textit{a priori} that it should better interact with the learning of a cosine metric than MSE due to the similar form between the two. This leads to our first hypothesis (H1):
\begin{hypo}
Learning a cosine distance along a representation with CRL gives better results than with MSE.
\end{hypo} 
Despite the possible issues, we hope that learning both tasks jointly should lead to the learning of more appropriate representations and thus to better results. This leads to our second hypothesis (H2):
\begin{hypo}
Jointly learning a metric and a representation with a sequence to sequence model gives better results than learning both separately.
\end{hypo}

\section{Experiments}
\label{exp}

\subsection{Experimental Setup}
\subsubsection{Dataset Presentation.}

\begin{figure}[h!]
\begin{subfigure}[b]{\textwidth}
\centering
\includegraphics[width=0.6\textwidth]{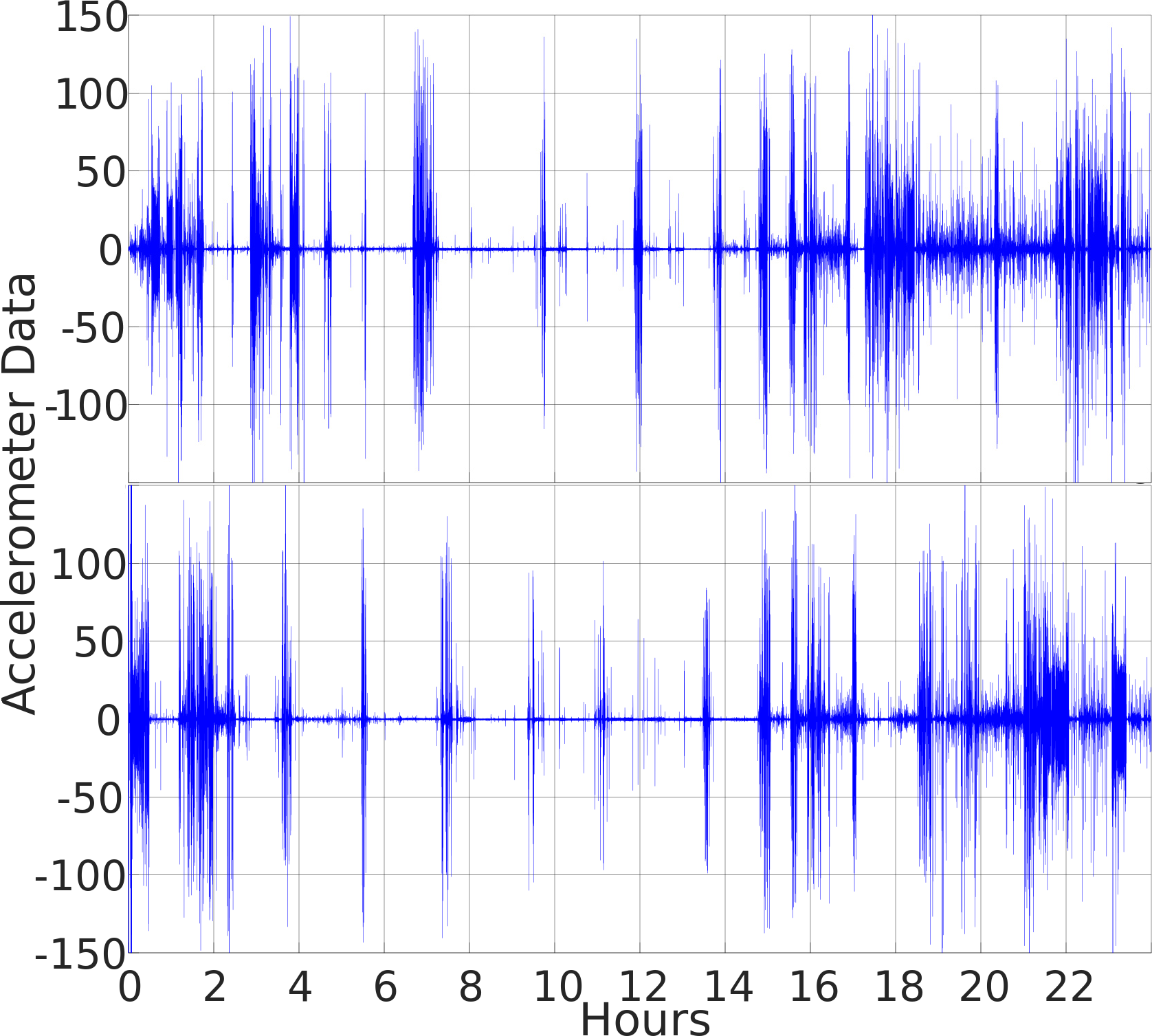}
\caption{2 days of accelerometer data.}
\end{subfigure}\\
\begin{subfigure}[b]{\textwidth}
\centering
\includegraphics[width=0.8\textwidth]{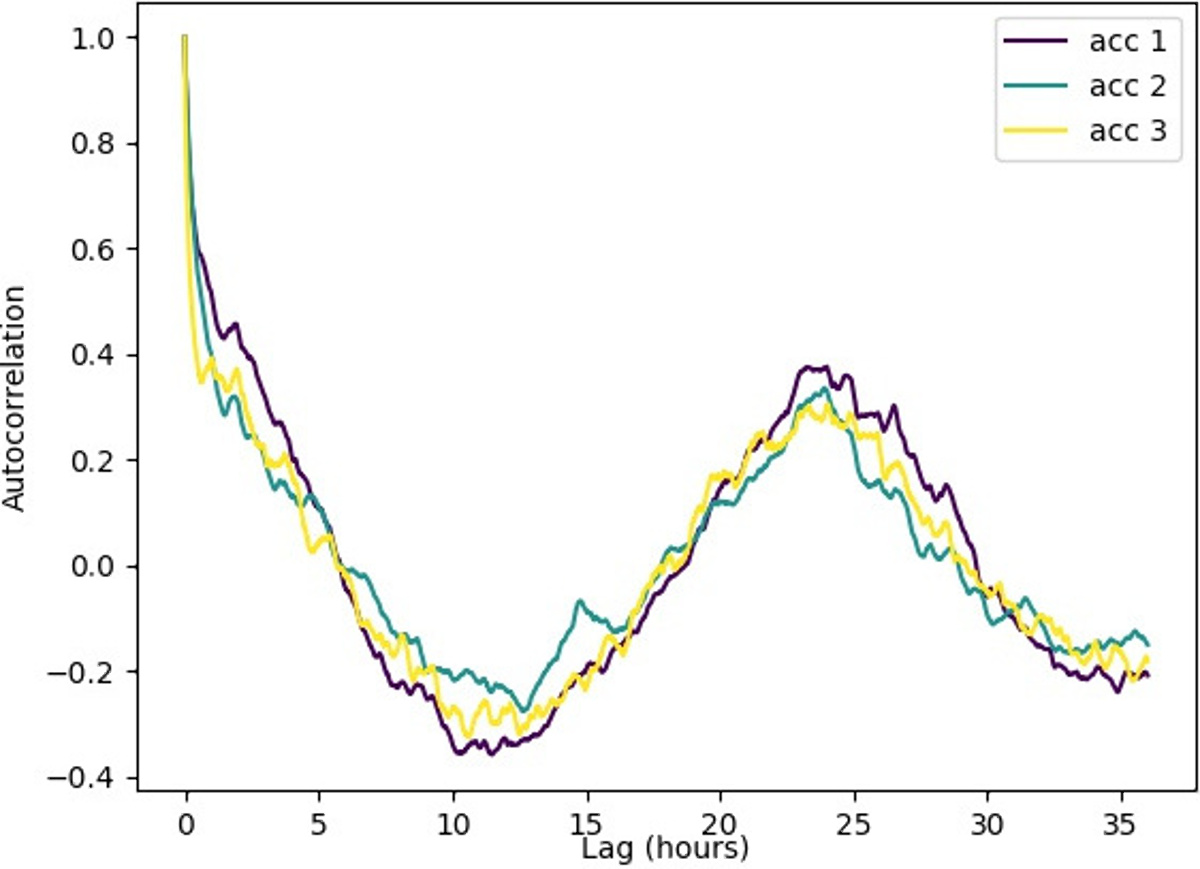}
\caption{Input signal autocorrelation for accelerometer data.}
\end{subfigure}
\caption{LTMM dataset used to evaluate routine modeling procedure.}
\label{data}
\end{figure}

Long-term unscripted data from wearable sensors are difficult to gather. The only dataset we found that could fit our requirements has been obtained by Weiss \al \cite{weiss2013does} and is called Long Term Movement Monitoring dataset (LTMM)\footnote{\url{https://www.physionet.org/physiobank/database/ltmm}}. This dataset contains recordings of 71 elderly people which have worn an accelerometer and a gyroscope during three days with no instructions. This dataset contains no labels. Fig \ref{data}.a presents two days of data coming from one axis of the accelerometer: similar profiles can be observed at similar moment. Fig \ref{data}.b presents the autocorrelation of the accelerometer signal: the maximum of 0.4 is reached for a phase of 24h. These figures show the interest of this dataset as the data show periodic nature while presenting major visual differences. That said, the definition of periodicity that our algorithm is made to achieve is stronger as it is based on a metric between extracted feature vectors, not just correlations of signal measurements. \\
To constitute our dataset, we selected in the original dataset a user who did not remove the sensor during the three days to avoid missing values. We set up a data augmentation process to artificially increase the quantity of data while preserving its characteristic structure. The dataset is sampled at 100 Hz and thus, to multiply the number of days by ten, each vector measurement at the same index modulo 10 will be affected to a new day (the order is respected). This new dataset has a sampling rate of 10 Hz which means that one hour of data is a sequence of size 36000, we consider only non overlapping sequences. Thus, to make the computation more tractable, polyphase filtering is applied to resample each sequence of one hour to a size of 100. Finally, equivalence constraints need to be defined in order to make similar and dissimilar pairs: two sequences of one hour, not from the same day but recorded at the same time are considered similar, all other combinations are considered dissimilar. This approach does not therefore require semantic labels. 

\subsubsection{Model Parameters and Training Details.}

We describe here the hyperparameters used to train the models. The autoencoders are constituted of one layer of 100 LSTM neurons for the encoder and the decoder. For the KISSME version, the encodings are then projected into a 50-dimensional space, and the distance matrix, which thus has also dimension 50, was updated with the closed-form every 30 epochs.
These parameters were determined after preliminary tests where deeper architectures and higher dimensional spaces were tested. Models are trained with 20 similar pairs for each time slot and the same total number of dissimilar pairs for a total of 960 training pairs coming from 12 different days of data. The training was stopped based on the loss computed on the validation set which contains three days of data i.e., 72 sequences. The testing set is composed of 15 days or 360 sequences. The data in the training set were rescaled between -1 and 1 and the same parameters were applied on the validation and testing sets. A learning rate of 0.001 was used and divided by 10 if the loss did not decrease anymore during 10 epochs. A batch size of 50, a margin of 1 for the contrastive loss and of 0.5 for the cosine loss were chosen. We also observed that changing to zero 30\% of the values of the training sequences sliglty improved the results as suggested in \cite{vincent2008extracting}.

\subsection{Experimental Results and Discussion}
Since the only available labels are time indications and to keep minimal supervision, the evaluation metrics rely on clustering. We report average values on 20 tests for 4 clustering evaluation metrics. \textit{Completeness} assesses if sequences produced at the same hour are in the same clusters. \textit{Silhouette} describes the cluster shapes, if they are dense and well-separated. \textit{Normalized Mutual Information} (NMI) is a classical metric for clustering and measures how two clustering assigments concur, the second being the time slots. \textit{Adjusted Mutual Information} (AMI) is the adjusted against chance version of NMI. A spectral clustering into 5 clusters is performed with the goal not to find the precise number of clusters maximizing the metrics but to choose a number which will make appear coherent and interpretable routines of the day, namely sleep moments, meals and other daily activities performed every day. Finally, to make our distances usable by the spectral clustering, they are converted to kernel functions. The following transformation was applied to the Euclidean, Mahalanobis and DTW distances: $\text{exp}(-\text{dist} \cdot \gamma)$ where $\gamma$ is the inverse of the length of an encoding vector (respectively number of features time length of sequence for DTW). $1$ was added to the cosine similarity so it becomes a kernel. 
\subsubsection{Evaluation of Cosine Reconstruction Loss.}

The performance of the CRL on LTMM is first evaluated. An experiment was performed by jointly training models for Euclidean or cosine distances with CRL or MSE. The results are reported in Table \ref{exp_crl}. An asterisk means the average results are significantly higher according to a Welch's test. The results demonstrate a significant improvement of the proposed CRL over MSE when trained with the cosine similarity for Completeness, NMI and AMI. For the Silhouette score, better results are obtained with the MSE. However, the standard deviations are large, and this improvement is thus not significant. With the Euclidean distance, the same improvement is not realized with a slight advantage of MSE over CRL. These results confirm our hypothesis H1 that it is more appropriate to learn a cosine distance with CRL. They also suggest a positive interaction between the two as the same effect could not be observed with the Euclidean distance. We then use CRL in the remaining of the paper.

\begin{table}
\begin{subtable}{.5\textwidth}
\centering
 \begin{tabular}{| c | c | c |}
 \hline
 \backslashbox{MeL}{RL} & \textbf{CRL} & \textbf{MSE}\\ \hline
 \textbf{Cosine} & \textbf{0.714}* $\pm$ 0.048&0.666 $\pm$ 0.066 \\ \hline
 \textbf{Euclidean} & 0.609 $\pm$ 0.042& 0.635 $\pm$ 0.064\\ \hline
 \end{tabular}
 \caption{Completeness}
 \end{subtable}%
 \begin{subtable}{.5\textwidth}
 \centering
 \begin{tabular}{| c | c | c |}
 \hline
 \backslashbox{MeL}{RL} & \textbf{CRL} & \textbf{MSE}\\ \hline
 \textbf{Cosine} & 0.618 $\pm$ 0.105 &\textbf{0.667} $\pm$ 0.144\\ \hline
 \textbf{Euclidean} & 0.402 $\pm$ 0.05&0.408 $\pm$ 0.042\\ \hline
 \end{tabular}
 \caption{Silhouette}
 \end{subtable}%
 
 \begin{subtable}{.5\textwidth}
 \centering
 \begin{tabular}{| c | c | c |}
 \hline
 \backslashbox{MeL}{RL} & \textbf{CRL} & \textbf{MSE}\\ \hline
 \textbf{Cosine} & \textbf{0.449}* $\pm$ 0.032&0.397 $\pm$ 0.040 \\ \hline
 \textbf{Euclidean} & 0.419 $\pm$ 0.033& 0.434 $\pm$ 0.047 \\ \hline
 \end{tabular}
 \caption{NMI}
 \end{subtable}%
 \begin{subtable}{.5\textwidth}
 \centering
 \begin{tabular}{| c | c | c |}
 \hline
		 \backslashbox{MeL}{RL} & \textbf{CRL} & \textbf{MSE}\\ \hline
 \textbf{Cosine} & \textbf{0.253}* $\pm$ 0.03&0.205 $\pm$ 0.033\\ \hline
 \textbf{Euclidean} & 0.255 $\pm$ 0.027&0.264 $\pm$ 0.038\\ \hline
 \end{tabular}
 \caption{AMI}
 \end{subtable}%

\caption{Evaluations of CRL and MSE on LTMM dataset.}
\label{exp_crl}
\end{table}
 
\subsubsection{Evaluation of the SS2S Architecture.}

Next, we investigated the benefit of the SS2S architecture over DTW and Siamese LSTM (SLSTM) \cite{mueller2016siamese} as well as the interest of jointly learning the encoder-decoder and the metric on the LTMM dataset. Results are presented in Table \ref{v1}. To test the DTW, the better radius was selected on the validation set and the spectral clustering was performed using DTW as kernel. Although Completeness, NMI and AMI are higher than every SS2S architectures except one, we observe a negative silhouette value which indicates a poor quality of the clustering and seems to confirm than indeed shaped based distances are not suitable for this type of data. Concerning the encoding architecture, SS2S gives overall better results than SLSTM and the best results are achieved by using the disjoint version of KISSME with a completeness of 0.983 and an NMI of 0.619. These results are not surprising as KISSME uses a parametric distance which can therefore be more adapted to the data. For the silhouette score, cosine distances performed best, i.e., they learned more compact and well-defined clusters. We also note that disjoint versions of the architectures performed better than the joint versions, thus invalidating our hypothesis H2. \\

To investigate the reasons of this difference which could be due to the autoencoder not being learned properly, Table \ref{vre} reports average best Reconstruction Errors on Validation set (REV). The lowest errors are systematically achieved when the encoder is learned alone before the metric therefore supporting the hypothesis that learning the metric prevents the autoencoder from being trained at its full potential. It explains why the joint learning does not perform best. For the CRL, results are closer than for MSE suggesting why this reconstruction loss is easier to learn jointly. \\

\begin{figure}[h!]
\centering
\begin{subfigure}[t]{\textwidth}
 \centering
 \includegraphics[width=0.8\textwidth]{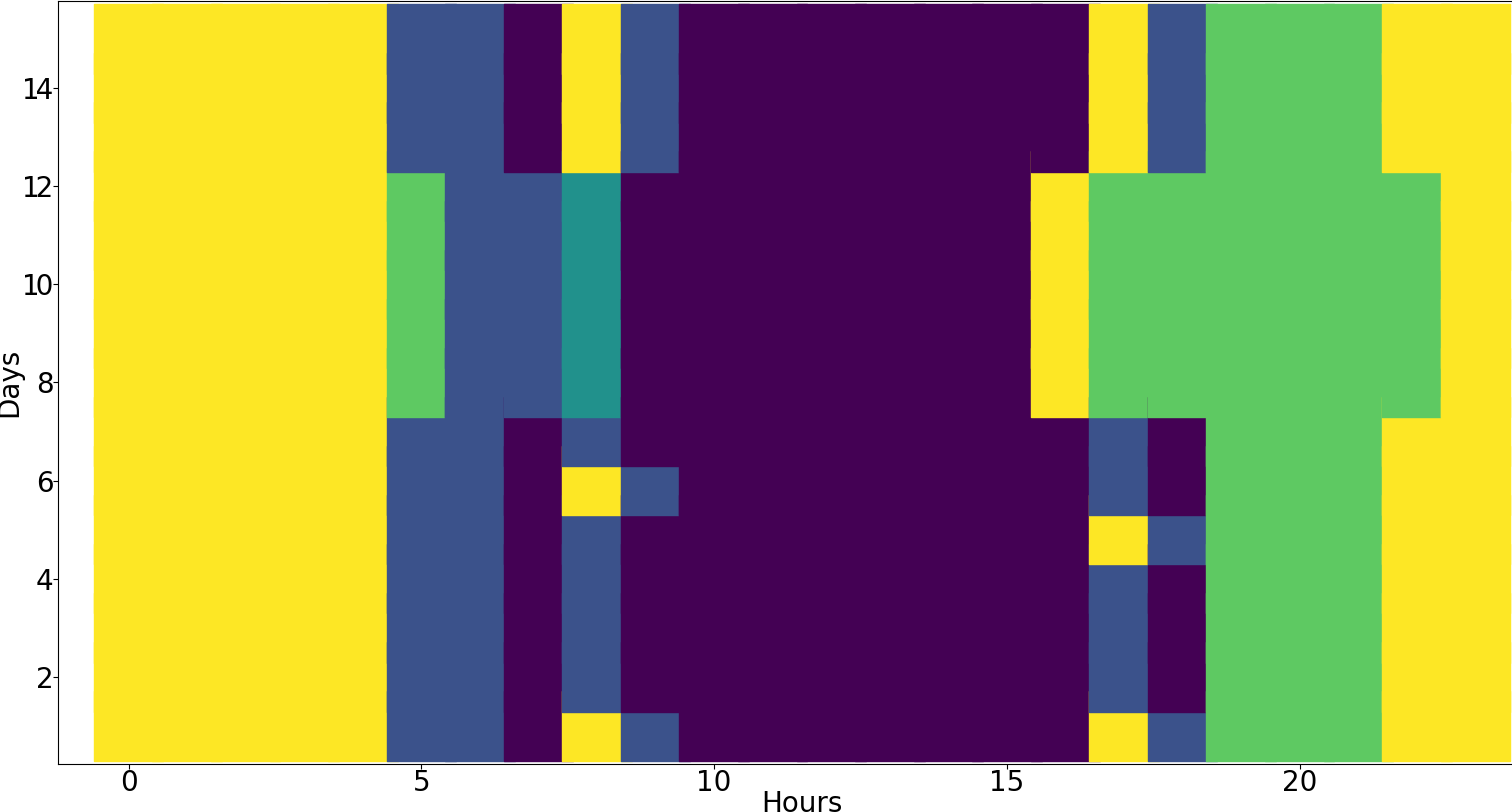}
 \caption{DTW \cite{salvador2007toward}.}
 \end{subfigure}\\
 \begin{subfigure}[t]{\textwidth}
 \centering
 \includegraphics[width=0.8\textwidth]{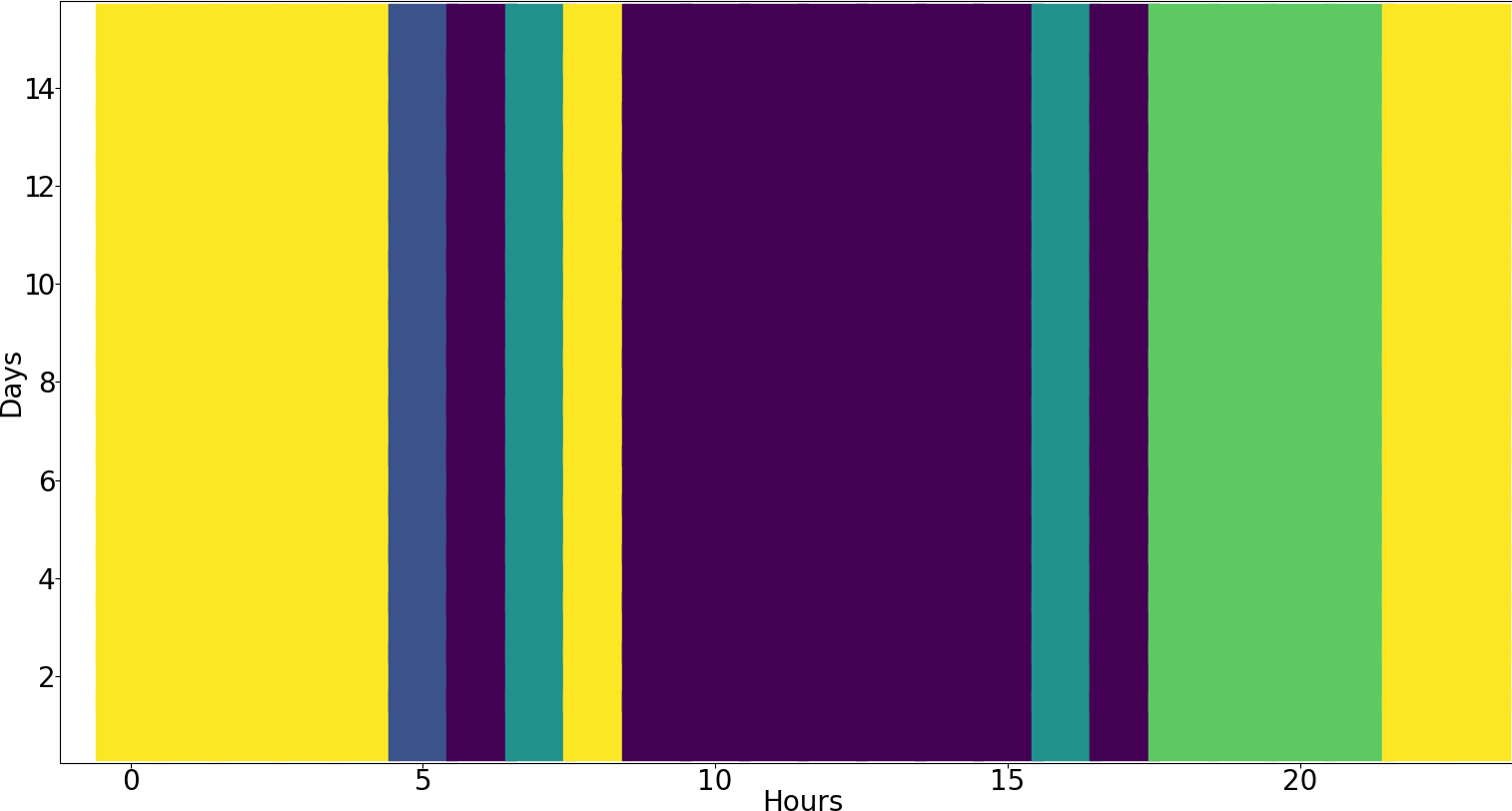}
 \caption{SS2S and KISSME, disjoint learning.}
 \end{subfigure}
 \caption{Examples of clustering obtained with our model on LTMM. 
 }
\label{fig_cluster}
\end{figure}

Finally, Fig. \ref{fig_cluster} shows clustering representations for two approaches: DTW and disjoint KISSME. The clusterings reflect the sequences of one hour that were found similar across the days on the testing set. If these sequences are at the same hour or cover the same time slots, we can argue it is a recurrent activity (or succession of activities) and therefore a routine. The disjoint KISSME version exhibits more coherent discrimination of routines, which, according to the 4 evaluation metrics reported was predictable. Several misclassified situations seem to appear for DTW which is coherent with the negative silhouette score. High regularities can be observed, and it is actually possible to make interpretations: yellow probably corresponds to sleeping moments and nights, and purple to activities during the day. Other clusters seem to correspond to activities at the evening or during meal time. Consequently, the SS2S architecture is able to learn a metric which cluster and produce a modeling of the daily routines of the person without labels. In this example, the clusters are coarse, the granularity of this analysis could be improved simply by working with sequences of half an hour or even shorter and produce more clusters.

\begin{table}
\centering
\resizebox{\textwidth}{!}{
\begin{tabular}{|c|c|c|c|c|c|c|c|}
\hline
\textbf{Metric} & \textbf{Model} & \textbf{Joint} & \textbf{Completeness} & \textbf{Silhouette} & \textbf{NMI} & \textbf{AMI} \\
\hline
DTW \cite{salvador2007toward} & x & x & 0.804 & -0.93 & 0.528 & 0.32 \\
\hline
Euclidean & SLSTM& x& 0.616 $\pm$ 0.032 & 0.427 $\pm$ 0.053& 0.414 $\pm$ 0.022 & 0.246 $\pm$ 0.019\\
\hline
Cosine& SLSTM& x& 0.617 $\pm$ 0.06 & 0.572 $\pm$ 0.143& 0.372 $\pm$ 0.052 & 0.192 $\pm$ 0.046\\
\hline 
Euclidean & SS2S & no &0.674 $\pm$ 0.04 & 0.528 $\pm$ 0.07& 0.458 $\pm$ 0.03 & 0.28  $\pm$ 0.027\\
\hline
Euclidean& SS2S & yes & 0.635 $\pm$ 0.064 & 0.408 $\pm$ 0.042 & 0.434 $\pm$ 0.047 &0.264 $\pm$ 0.038 \\
\hline
Cosine& SS2S & no &0.71 $\pm$ 0.05& \textbf{0.756*} $\pm$ 0.089& 0.467 $\pm$ 0.028& 0.275 $\pm$ 0.024\\
\hline
Cosine& SS2S & yes & 0.714 $\pm$ 0.048 & 0.618 $\pm$ 0.105 & 0.449 $\pm$ 0.032 & 0.253 $\pm$ 0.03 \\
\hline
KISSME & SS2S & no & \textbf{0.983*} $\pm$ 0.016& 0.439 $\pm$ 0.077& \textbf{0.619*} $\pm$ 0.035& \textbf{0.363*} $\pm$ 0.046\\
\hline
KISSME & SS2S & yes & 0.667 $\pm$ 0.021& 0.316 $\pm$ 0.039& 0.446 $\pm$ 0.012& 0.266 $\pm$ 0.012 \\
\hline
\end{tabular}}
\caption{Evaluations on LTMM dataset of the SS2S architecture (x means non applicable).}
\label{v1}
\end{table}

\begin{table}
\centering
\begin{subtable}[b]{.5\textwidth}
\centering
\begin{tabular}{|c|c|}
\hline
\textbf{Metric} & \textbf{REV} \\
\hline
Euclidean & 0.707 $\pm$ 0.112\\
\hline
KISSME & 0.736 $\pm$ 0.099\\
\hline
Disjoint & \textbf{0.55}* $\pm$ 0.083\\
\hline
\end{tabular}
\caption{MSE}
\end{subtable}%
\begin{subtable}[b]{.5\textwidth}
\centering
\begin{tabular}{|c|c|}
\hline
\textbf{Metric} & \textbf{REV} \\
\hline
Cosine & 0.339 $\pm$ 0.036\\
\hline
Disjoint & \textbf{0.298}* $\pm$ 0.03\\
\hline
\end{tabular}
\caption{CRL}
\end{subtable}
\caption{Average reconstruction errors on the validation set of LTMM.}
\label{vre}
\end{table}

\section{Conclusions and perspectives}
We presented a metric learning model to cluster routines in the daily behavior of individuals. By defining routines as almost-periodic functions, we have been able to study them in a metric learning framework. We thus proposed an approach which combines metric learning and representation learning of sequences. Our proposed architecture relies on no labels and is learned only from time slots. A new reconstruction loss was also proposed to be learned jointly with a cosine metric and it showed better results than MSE in this case. Our SS2S architecture with KISSME and disjoint learning process achieved stimulating results with 0.983 of completeness and 0.619 of NMI. A visual evaluation analysis allows to interpret the recurrent behaviors discovered by the architecture. However, these results invalidate in this case our second hypothesis that combining metric learning and sequence to sequence learning would give better results. \\
In the future, we will investigate more deeply joint learning of representations and metrics. Several architecture improvements could also be made, for examples: work with triplets instead of pairs, replace the LSTM with a convolutionnal neural network \cite{gehring2017convolutional} or an echo states network \cite{jaeger2001echo}. This last approach works quite differently from a normal neural network and would require subsequent modifications of the architecture. Finally, we will study in further details the link between almost-periodic functions and metric learning. \\

\bibliographystyle{splncs03}
\bibliography{biblio}

\end{document}